\documentclass[10pt, conference, compsocconf, letterpaper]{IEEEtran}
\IEEEoverridecommandlockouts

\usepackage{amsthm}
\usepackage{cite}
\usepackage{amsmath,amssymb,amsfonts}
\usepackage[ruled,vlined,linesnumbered]{algorithm2e}
\usepackage{algpseudocode} 
\usepackage{graphicx}
\usepackage{textcomp}
\usepackage{subfig}
\usepackage{multirow}
\usepackage{tikz}
\usetikzlibrary{automata, positioning}

 \def\BibTeX{{\rm B\kern-.05em{\sc i\kern-.025em b}\kern-.08em
    T\kern-.1667em\lower.7ex\hbox{E}\kern-.125emX}}
\usepackage{enumitem}
\usepackage{colortbl}
\usepackage{soul}
\usepackage{graphicx} 
\usepackage[colorlinks=true, allcolors=blue]{hyperref}%

\newlist{inlineroman}{enumerate*}{1}
\setlist[inlineroman]{itemjoin*={{, and }},afterlabel=~,label=\roman*.}

\newlist{Inlineroman}{enumerate*}{1}
\setlist[Inlineroman]{itemjoin*={{, and }},afterlabel=~,label=\Roman*.}

\author{Jumman Hossain$^{1}$, Abu-Zaher Faridee$^{1,2}$, Derrik Asher$^{3}$, Jade Freeman$^{3}$, Theron Trout$^{4}$, Timothy Gregory$^{3}$ \\ and Nirmalya Roy$^{1}$%
\thanks{*This work has been supported by U.S. Army Grant \texttt{\#W911NF2120076}.}%
\thanks{$^{1}$Dept. of Information Systems, University of Maryland, Baltimore County, USA. {\tt\scriptsize \{jumman.hossain, faridee1, nroy\}@umbc.edu}.}%
\thanks{$^{2}$Amazon Inc., USA. {\tt\scriptsize abufari@amazon.com}.}%
\thanks{$^{3}$DEVCOM Army Research Lab, USA.}%
\thanks{$^{4}$Stormfish Scientific Corporation, {\tt\scriptsize theron.trout@stormfish-sci.com}.}%
}

\begin{document}

\setlist[enumerate]{nosep}
\setlist[itemize]{nosep}
\definecolor{mintgreen}{rgb}{0.6, 1.0, 0.6}
\definecolor{pastelviolet}{rgb}{0.8, 0.6, 0.79}
\definecolor{peridot}{rgb}{0.9, 0.89, 0.0}
\definecolor{richbrilliantlavender}{rgb}{0.95, 0.65, 1.0}
\definecolor{robineggblue}{rgb}{0.0, 0.8, 0.8}

\definecolor{green}{rgb}{0.1,0.1,0.1}
\newcommand{\done}{\cellcolor{teal}done}

\title{\textit{QuasiNav}: Asymmetric Cost-Aware Navigation Planning with Constrained Quasimetric Reinforcement Learning}


\maketitle
\begin{abstract}

Autonomous navigation in unstructured outdoor environments is inherently challenging due to the presence of asymmetric traversal costs, such as varying energy expenditures for uphill versus downhill movement. Traditional reinforcement learning methods often assume symmetric costs, which can lead to suboptimal navigation paths and increased safety risks in real-world scenarios. In this paper, we introduce \textit{QuasiNav}, a novel reinforcement learning framework that integrates quasimetric embeddings to explicitly model asymmetric costs and guide efficient, safe navigation. \textit{QuasiNav} formulates the navigation problem as a constrained Markov decision process (CMDP) and employs quasimetric embeddings to capture directionally dependent costs, allowing for a more accurate representation of the terrain. This approach is combined with adaptive constraint tightening within a constrained policy optimization framework to dynamically enforce safety constraints during learning. We validate \textit{QuasiNav} across three challenging navigation scenarios—undulating terrains, asymmetric hill traversal, and directionally dependent terrain traversal—demonstrating its effectiveness in both simulated and real-world environments. Experimental results show that \textit{QuasiNav} significantly outperforms conventional methods, achieving higher success rates, improved energy efficiency, and better adherence to safety constraints.
\end{abstract}

\section{Introduction}
Autonomous navigation in unstructured outdoor environments presents significant challenges due to asymmetric traversal costs that vary with direction, such as differing energy requirements for uphill versus downhill movement ~\cite{eder2023predicting, wei2023predicting}. These asymmetric costs are often ignored by traditional reinforcement learning (RL) methods, which typically assume symmetric cost structures, leading to suboptimal and sometimes unsafe navigation paths ~\cite{gupta2017cognitive, levine2016end}. Efficient and safe navigation in such environments requires a more sophisticated approach that accurately models these directional dependencies, ensuring that the robot can make informed decisions that optimize both path efficiency and safety.

\begin{figure}
    \centering
    \includegraphics[width=\linewidth]{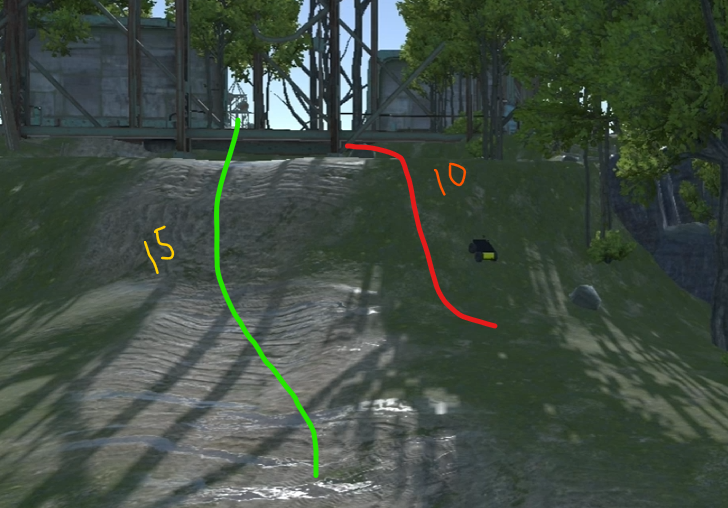}
    \caption{An illustration of asymmetric traversal costs during navigation. The green path ($15^\circ$ slope) represents a gentler, more energy-efficient route, while the red path ($10^\circ$ slope) is steeper and potentially riskier. This scenario demonstrates how \textit{QuasiNav}'s quasimetric embedding can capture the directional dependence of terrain traversability, favoring the longer but safer and more efficient green path over the shorter but more challenging red path.}
    \label{fig:QCPO_navigation_planning}
\end{figure}

Recent advances in RL and inverse reinforcement learning (IRL) have attempted to address these challenges by incorporating expert demonstrations and cost maps that better reflect complex terrain interactions \cite{gupta2017cognitive}. However, these methods often struggle with scalability and adaptability in new environments, especially where asymmetric costs are prominent \cite{levine2016end}. Other approaches, such as semantic mapping, improve environmental understanding but frequently rely on static or predefined rules that do not dynamically adjust to changing conditions, limiting their effectiveness in real-world scenarios \cite{papandreou2015semantic, milioto2019rangenet}.

To overcome these limitations, we propose \textit{QuasiNav}, a novel reinforcement learning framework that integrates quasimetric embeddings to explicitly model asymmetric traversal costs. Fig. \ref{fig:QCPO_navigation_planning} illustrates an example of asymmetric traversal costs in off-road navigation, demonstrating the need for our quasimetric approach. Quasimetrics ~\cite{wang2022learning, liu2023metric, pitis2020inductive} differ from traditional metric spaces by allowing directional dependencies, providing a more accurate representation of terrain costs that vary based on the direction of movement. By embedding these quasimetrics into a constrained policy optimization framework, \textit{QuasiNav} is able to navigate complex terrains by selecting paths that minimize traversal costs while dynamically enforcing safety constraints through adaptive constraint tightening \cite{achiam2017constrained, wang2022quasimetric}.



Our experimental results show that \textit{QuasiNav} achieves substantial improvements over existing navigation methods by effectively handling asymmetric traversal costs and dynamically adapting to varying terrain conditions. \textit{QuasiNav} consistently achieves higher success rates, optimizes energy consumption by reducing unnecessary elevation changes, and enhances overall safety compliance in complex unstructured environments.

\textbf{Main Contributions}: \begin{itemize} \item \textbf{Quasimetric Embeddings for Asymmetric Costs}: We introduce a novel approach to model directionally dependent traversal costs in unstructured outdoor environments using quasimetric embeddings. This method captures terrain-specific asymmetries such as slope differences, surface friction, and distance-based asymmetry, enabling more accurate representation of complex navigation scenarios.

\item \textbf{Constrained Policy Optimization with Adaptive Tightening}: Our framework integrates adaptive constraint tightening within a CMDP, ensuring that learned policies are both efficient and safe under varying terrain conditions. 

\item \textbf{Energy-Aware Navigation}: We propose a comprehensive energy consumption model that accounts for terrain-specific factors such as slope, surface friction, and directional dependencies. This model is integrated into our reward function and evaluation metrics, allowing for a more accurate assessment of navigation efficiency in terms of energy expenditure. Our experimental results demonstrate significant improvements in energy efficiency, with \textit{QuasiNav} achieving up to 13.6\% reduction in energy consumption compared to baseline methods.

\item \textbf{Comprehensive Evaluation in Complex Scenarios}: We validate \textit{QuasiNav} across three challenging navigation scenarios—undulating terrains, asymmetric hill traversal, and directionally dependent terrain traversal—demonstrating its effectiveness in both simulated and real-world environments. Our results show that \textit{QuasiNav} significantly outperforms conventional methods, achieving higher success rates, improved energy efficiency, and better adherence to safety constraints.

\end{itemize}
\section{Related Work}

\subsection{Asymmetric Cost Modeling in Navigation}

Modeling asymmetric traversal costs has become a critical aspect of navigation in challenging environments. Recent work by Suh and Oh \cite{suh2012cost} introduced a cost-aware path planning algorithm for mobile robots, emphasizing the importance of incorporating cost variations into planning to improve energy efficiency and safety. Similarly, DEM-AIA, an asymmetric inclination-aware trajectory planner, specifically addresses the challenges of off-road navigation by using digital elevation models to account for uneven terrains \cite{dem-aia2023}. These studies highlight the importance of directionally dependent cost modeling, but often lack a direct integration with real-time safety constraints, which is crucial for autonomous systems operating in dynamic environments.

\subsection{Constrained Policy Optimization for Safe Navigation}

Constrained Markov Decision Processes (CMDPs) have been employed to incorporate safety constraints into navigation policies. Constrained Policy Optimization (CPO) by Achiam et al. \cite{achiam2017constrained} represents a significant advance in ensuring that RL agents adhere to safety boundaries during learning. However, these methods typically assume symmetric traversal costs, which can be suboptimal for environments with directional dependencies such as slopes or asymmetric obstacles. Recent work has sought to extend these principles by adapting CMDPs to handle asymmetric costs, but challenges remain in fully integrating these approaches with directionally dependent navigation requirements \cite{ma2023safe}.

\subsection{Quasimetric Reinforcement Learning (QRL)}

Quasimetric Reinforcement Learning (QRL) introduces a novel approach to handling directionally dependent traversal costs, essential for optimizing paths in asymmetric environments. QRL methods, such as those by Wang et al. \cite{wang2022quasimetric, wang2023optimal}, utilize quasimetric models to represent costs that vary based on direction, enhancing path planning in controlled simulations like MountainCar and other goal-reaching benchmarks ~\cite{tongzhouw2023qrl}. However, existing QRL approaches have primarily focused on idealized environments and have not been extended to unstructured outdoor settings where real-world complexities, including safety constraints, are critical. This gap highlights the need for approaches like \textit{QuasiNav} that integrate quasimetric embeddings ~\cite{tongzhouw2022iqe,memoli2018quasimetric} with robust safety mechanisms.

\textit{QuasiNav} extends the principles of QRL by integrating quasimetric embeddings ~\cite{wang2022improved} into a constrained policy optimization framework, explicitly modeling asymmetric traversal costs and dynamically enforcing safety constraints through adaptive constraint tightening. This approach allows \textit{QuasiNav} to navigate complex terrains with dynamically feasible paths, optimizing both efficiency and safety. Our method is validated across diverse scenarios, demonstrating significant improvements in navigation performance over traditional RL methods and recent QRL approaches. To the best of our knowledge, this is the first work to leverage quasimetric embeddings for learning navigation policies in unstructured outdoor environments while explicitly considering safety constraints.
\section{Background and Problem Formulation}

\subsection{Problem Definition}
\label{subsec:problem_definition}
The outdoor environment consists of different types of terrain, each with its own characteristics and traversal costs. For example, the robot may encounter grass, dirt, gravel, and obstacles like rocks and trees. The goal is to navigate from a starting position to a target location while minimizing the total cost incurred during the traversal. 

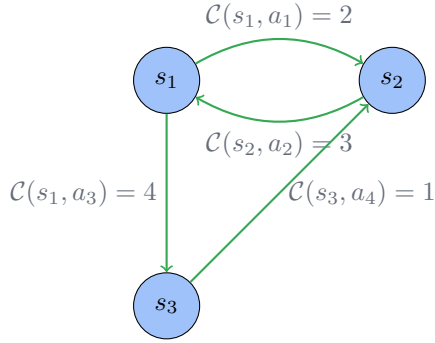
\begin{figure}[h]
    \centering
    \begin{tikzpicture}[node distance=3cm]
        \definecolor{statecolor}{RGB}{66, 133, 244} 
        \definecolor{arrowcolor}{RGB}{52, 168, 83} 
        \definecolor{textcolor}{RGB}{101, 109, 120} 

        \node[state, fill=statecolor!50, text=black] (A) {$s_1$};
        \node[state, fill=statecolor!50, text=black, right of=A] (B) {$s_2$};
        \node[state, fill=statecolor!50, text=black, below of=A] (C) {$s_3$};

        \draw[->, thick, arrowcolor] (A) to[bend left] node[above, text=textcolor] {$\mathcal{C}(s_1, a_1) = 2$} (B);
        \draw[->, thick, arrowcolor] (B) to[bend left] node[below, text=textcolor] {$\mathcal{C}(s_2, a_2) = 3$} (A);
        \draw[->, thick, arrowcolor] (A) -- node[left, text=textcolor] {$\mathcal{C}(s_1, a_3) = 4$} (C);
        \draw[->, thick, arrowcolor] (C) -- node[right, text=textcolor] {$\mathcal{C}(s_3, a_4) = 1$} (B);
    \end{tikzpicture}
    \caption{A sample state transition diagram with associated costs.}
    \label{fig:state_transition}
\end{figure}

For example, in fig. \ref{fig:state_transition}, we have three states: $s_1$, $s_2$, and $s_3$, representing different positions in the off-road environment. The arrows between the states indicate the actions $a_1$, $a_2$, $a_3$, and $a_4$, which allow the robot to move from one state to another. The costs associated with each state-action pair are denoted by $\mathcal{C}(s_i, a_j)$.
The quasimetric property of the cost function is evident in this example:

\textbf{Triangle Inequality:} The cost of going directly from $s_1$ to $s_2$ is 2, which is less than or equal to the cost of going from $s_1$ to $s_3$ (cost = 4) and then from $s_3$ to $s_2$ (cost = 1), i.e., $\mathcal{C}(s_1, a_1) \leq \mathcal{C}(s_1, a_3) + \mathcal{C}(s_3, a_4)$.

\textbf{Asymmetry:} The cost of going from $s_1$ to $s_2$ (cost = 2) is not necessarily equal to the cost of going from $s_2$ to $s_1$ (cost = 3), i.e., $\mathcal{C}(s_1, a_1) \neq \mathcal{C}(s_2, a_2)$.

The asymmetric cost function in \textit{QuasiNav} captures the directional dependencies inherent in off-road navigation. For instance, uphill travel typically costs more than downhill, and terrain difficulty can vary based on approach direction. By employing a quasimetric cost function, our approach accurately models these asymmetries, enabling more informed and context-aware navigation policies in unstructured outdoor environments.

\subsection{Problem Formulation}
\label{subsec:problem_formulation}

We formulate the navigation problem as a constrained Markov decision process (CMDP) to minimize traversal costs while adhering to safety constraints.
The CMDP is defined by the tuple $(\mathcal{S}, \mathcal{A}, \mathcal{P}, \mathcal{C}, \mathcal{G}, \mu, \gamma)$, where:
\begin{itemize}
\item $\mathcal{S} \subseteq \mathbb{R}^n$ is the state space, which includes the robot's position, orientation, and velocity, as well as any relevant environmental features.
\item $\mathcal{A} \subseteq \mathbb{R}^m$ is the action space, which corresponds to the robot's control inputs (e.g., linear and angular velocities).
\item $\mathcal{P}: \mathcal{S} \times \mathcal{A} \times \mathcal{S} \to [0, 1]$ is the state transition probability distribution.
\item $\mathcal{C}: \mathcal{S} \times \mathcal{A} \to \mathbb{R}_{\geq 0}$ is the cost function, which assigns a non-negative cost to each state-action pair. In our setting, the cost function is assumed to be quasimetric, meaning that it satisfies the triangle inequality but may be asymmetric.
\item $\mathcal{G}: \mathcal{S} \to {0, 1}$ is the goal indicator function, which evaluates to 1 if the state is a goal state and 0 otherwise.
\item $\mu: \mathcal{S} \to [0, 1]$ is the initial state distribution.
\item $\gamma \in (0, 1)$ is the discount factor, which prioritizes near-term costs over long-term costs.
\end{itemize}
In addition to the standard CMDP components, we also consider a set of safety constraints $\mathcal{S}_{\text{safe}} \subseteq \mathcal{S}$, which define the set of allowable states that the robot must remain within to ensure safe operation. The goal is to learn a policy $\pi: \mathcal{S} \to \mathcal{A}$ that minimizes the expected cumulative cost while satisfying the safety constraints:
\begin{equation}
\begin{aligned}
\min_{\pi} J(\pi) &= \mathbb{E}_{\tau \sim \pi} \left[\sum_{t=0}^{\infty} \gamma^t C(s_t, a_t)\right] 
 \text{s.t.} \quad &s_t \in S_{\text{safe}} \; \forall t
\end{aligned}
\end{equation}
where $\tau = (s_0, a_0, s_1, a_1, \ldots)$ denotes a trajectory generated by following policy $\pi$ starting from an initial state $s_0 \sim \mu$.

\section{Methodology}
\label{sec:method}

\begin{figure}[h]
    \centering
    \includegraphics[width=\linewidth]{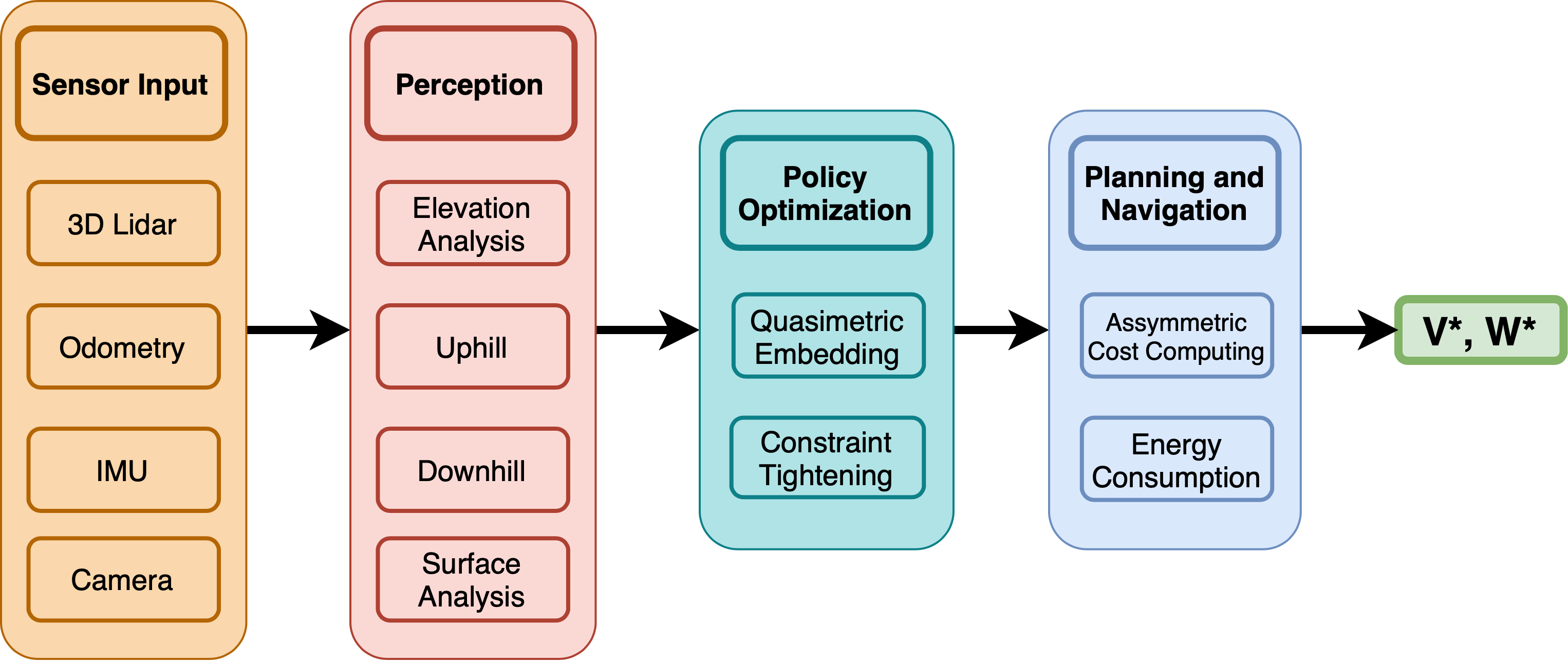}
    \caption{Overview of \textit{QuasiNav} System Architecture.} 
    \label{fig:architecture}
\end{figure}

In this section, we explain the major stages of our approach. Fig. \ref{fig:architecture} shows how different modules in our method are connected. 

\subsection{Feature Extraction from Environment}

\textit{QuasiNav} leverages a rich set of terrain features to accurately represent unstructured outdoor environments. Building on recent works TERP ~\cite{weerakoon2022terp} and GANav ~\cite{guan2022ga}, we extract a comprehensive feature vector $f(s)$ for each state $s$:
\begin{equation}
    f(s) = [z, \nabla z, \vec{n}, t, r, \mu, \rho, \kappa, \vec{d}]
\end{equation} where $z$ is the elevation, $\nabla z$ is the gradient, $\vec{n}$ is the surface normal, $t$ is the terrain class (obtained via efficient semantic segmentation), $r$ is the roughness, $\mu$ is the estimated friction coefficient, $\rho$ is the obstacle density, $\kappa$ is the path curvature, and $\vec{d}$ is the goal direction. These features are derived from fused LiDAR, RGB camera, and IMU data. The quasimetric embedding function $e_\psi(s,a)$ takes $f(s)$ and action $a$ as input, enabling it to learn a nuanced representation of asymmetric traversal costs in complex terrains.

\subsection{Asymmetric Cost Calculation}
 To model the asymmetries, we define the traversal cost $C(s, a)$ between states as a quasimetric function, which allows for asymmetric costs between transitions. The cost function $C(s, a)$ is calculated based on the robot’s motion and terrain properties. Specifically, we consider:
\begin{itemize}
    \item \textbf{Slope Asymmetry}: The cost of moving uphill is typically greater than downhill due to increased energy consumption. We model this as:
    \[
    C(s, a) = \begin{cases}
    \alpha \cdot |\Delta z| & \text{if} \ \Delta z > 0 \ (\text{uphill}) \\
    \beta \cdot |\Delta z| & \text{if} \ \Delta z \leq 0 \ (\text{downhill})
    \end{cases}
    \]
    where $\Delta z$ represents the change in elevation, and $\alpha > \beta$ reflect the higher cost of uphill traversal.

    \item \textbf{Surface Friction}: Different terrain types (e.g., gravel, grass, mud) have varying friction coefficients. The cost of traversing a terrain type is modeled as:
    \[
    C_{\text{terrain}}(s) = \lambda \cdot f_{\text{terrain}}(s)
    \]
    where $\lambda$ is a scaling factor, and $f_{\text{terrain}}(s)$ denotes the friction coefficient of the terrain at state $s$.

    \item \textbf{Distance-Based Asymmetry}: The directional cost of traveling between two points can differ based on environmental factors such as wind or path difficulty. This is incorporated as:
    \[
    C(s, s') = d(s, s') \cdot g(s, s')
    \]
    where $d(s, s')$ is the Euclidean distance between states and $g(s, s')$ represents a direction-dependent weight function.
\end{itemize}

\subsection{Quasimetric Embedding and Asymmetric Metric}

To model the asymmetric traversal costs encountered in unstructured environments, we utilize quasimetric embeddings. Unlike traditional metrics, where the distance between two points is symmetric, quasimetric spaces allow for directional dependence, making them ideal for capturing costs that vary based on traversal direction (e.g., uphill vs. downhill).

We define the quasimetric embedding $e_\phi(s, a)$, which maps state-action pairs into a high-dimensional space. The traversal cost between two state-action pairs $(s, a)$ and $(s', a')$ is represented as an asymmetric distance:
\[
d_H((s, a), (s', a')) = \| e_\phi(s, a) - e_\phi(s', a') \|_H
\]
where $\| \cdot \|_H$ is an asymmetric norm given by 
\begin{equation}
\|v\|_H = w^T \max(v, 0) + (1-w)^T \max(-v, 0)
\end{equation} This embedding captures terrain-dependent and directionally asymmetric traversal costs, such as increased effort for uphill movement or terrain friction differences.  For instance, in hill traversal, our embedding might yield \begin{equation}
\begin{aligned}
d_H((s_\text{bottom}, a_\text{climb}), (s_\text{top}, a_\text{null})) > \\ d_H((s_\text{top}, a_\text{descend}), (s_\text{bottom}, a_\text{null}))
\end{aligned}
\end{equation} capturing the higher energy cost and potential risk of climbing versus descending. The quasimetric distance $d_H$ is incorporated into the learning process by guiding the policy to favor actions that minimize total cost, while respecting the asymmetric nature of terrain traversal. The quasimetric embedding function is learned by minimizing a contrastive loss that pulls similar state-action pairs closer together and pushes dissimilar pairs farther apart. This loss is defined as:

\begin{equation}
\begin{aligned}
L(e_\phi) = \sum_{(s,a,s',a')} \Big( \| e_\phi(s, a) - e_\phi(s', a') \|_H^2 + \\
\max(0, m - d_H(s, a, s'', a'')) \Big)
\end{aligned}
\end{equation} where $(s'', a'')$ is a randomly sampled negative example, $d_H$ is the learned quasimetric distance, and $m$ is a margin hyperparameter.


\subsection{Constrained Policy Optimization}

We employ Constrained Policy Optimization (CPO) \cite{achiam2017constrained} to ensure that the robot's navigation policy adheres to safety constraints while minimizing traversal costs. The quasimetric embeddings $e_\phi(s, a)$ are integrated into the policy optimization process, allowing the robot to account for asymmetric costs during planning. We parameterize the policy $\pi_{\theta}$ using a neural network with parameters $\theta$. At each iteration, we sample a batch of trajectories ${\tau_i}_{i=1}^N$ using the current policy and compute the policy gradient:
\begin{equation}
g = \frac{1}{N} \sum_{i=1}^N \sum_{t=0}^{T} \nabla_{\theta} \log \pi_{\theta}(a_{i,t} | s_{i,t}) \hat{A}_{i,t},
\end{equation}
where $\hat{A}_{i,t}$ is an estimate of the advantage function at timestep $t$ of trajectory $i$, given by:
\begin{equation}
\hat{A}{i,t} = \sum{t'=t}^{T} \gamma^{t'-t} \mathcal{C}(s_{i,t'}, a_{i,t'}) - V_{\phi}(s_{i,t}).
\end{equation}
Here, $V_{\phi}(s) = \min_{a \in \mathcal{A}} \mathcal{C}(s, a) + \gamma \mathbb{E}{s' \sim \mathcal{P}(\cdot|s, a)}[V{\phi}(s')]$ is the value function estimated using the quasimetric embedding $\phi$.
To ensure that the policy update satisfies the safety constraints, we solve the following constrained optimization problem with the conjugate gradient method:
\begin{equation}
\begin{aligned}
\theta_{k+1} = \arg\min_{\theta} \quad & g^T (\theta - \theta_k) \
\textrm{s.t.}\\ \quad & \frac{1}{N} \sum_{i=1}^N \sum_{t=0}^{T} \mathbb{I}[s_{i,t} \notin \mathcal{S}_{\text{safe}}] \leq \delta,
\end{aligned}
\end{equation}
where $\delta \in (0, 1)$ is a constraint violation tolerance. 
At each iteration, the policy $\pi_\theta$ is updated by maximizing the expected cumulative reward, subject to safety constraints:
\[
\max_{\theta} \mathbb{E}_{\pi_\theta} \left[ \sum_{t=0}^{T} \gamma^t r(s_t, a_t) \right] \quad \text{subject to} \quad \mathbb{E}_{\pi_\theta} \left[ C(s_t, a_t) \right] \leq \delta
\]
where $\delta$ is the maximum allowable cost for safe navigation. The expected cumulative reward incorporates the asymmetric costs learned through quasimetric embeddings, ensuring that the robot prefers efficient and safe paths while respecting directional traversal difficulties. CPO uses a trust region method to update the policy within a safe range, ensuring that updates do not violate safety constraints. The overall process is detailed in Algorithm~\ref{alg:quasinav}




\begin{algorithm}
\caption{QuasiNav - Asymmetric Cost-Aware Navigation with QRL}
\label{alg:quasinav}
\KwIn{State space $\mathcal{S}$, action space $\mathcal{A}$, initial state $s_0$, goal state $g$, cost function $\mathcal{C}$, safety constraints $\mathcal{G}$, discount factor $\gamma$}
\KwOut{Optimized policy $\pi$}

Initialize policy $\pi$, quasimetric embedding function $e_\psi$, and environment model\;
Extract features $f(s) = [z, \nabla z, \vec{n}, t, r, \mu, \rho, \kappa, \vec{d}]$ from sensor data\;

\While{not converged}{
    \ForEach{episode}{
        $s \gets s_0$\;
        \While{$s \neq g$}{
            Compute $e_\psi(s, a)$ using current state $s$ and action $a$\;
            Evaluate asymmetric traversal costs using $e_\psi(s, a)$\;
            Update policy $\pi$ using constrained policy optimization (CMDP)\;
            Apply adaptive constraint tightening to enforce safety constraints $\mathcal{G}$\;
            $s \gets$ Execute action $a$ selected by policy $\pi$\;
        }
    }
    Update $e_\psi$ and $\pi$ with collected transitions $(s, a, s', a')$\;
}
\Return Optimized policy $\pi$\;
\end{algorithm}



\textbf{Proposition 1 (Optimality of QuasiNav Policy):} 
\textit{Let $e_\phi(s,a)$ be the optimal quasimetric embedding learned by QuasiNav for a deterministic Markov decision process (MDP) with $\gamma = 1$, and let $G$ be the graph representing all state-action pairs. If the optimal value function $V^*(s,s')$ is recovered for short-horizon goals $s'$, then finding the minimum-cost path in $G$ with edge weights $-V^*(s,s')$ recovers the optimal constrained policy $\pi^*$, which maximizes $V^*(s,g)$ while satisfying safety constraints.}

\begin{proof}
The Bellman equation for the constrained MDP is given by:\begin{equation}
V^*(s,g) = \max_a [-C(s,a,s') + V^*(s',g)]
\label{eq:bellman}
\end{equation}
subject to $C(s,a,s') \leq \delta$
where $C(s,a,s')$ is the cost function and $\delta$ is the safety constraint threshold. In QuasiNav, we define the cost function using the quasimetric embedding: \begin{equation}
C(s,a,s') = \|e_\phi(s,a) - e_\phi(s',a')\|_H
\label{eq:cost}
\end{equation}
where $\|\cdot\|_H$ is the asymmetric norm defined in Equation (3). Substituting \eqref{eq:cost} into \eqref{eq:bellman}, we get:
\begin{equation}
V^*(s,g) = \max_a [-\|e_\phi(s,a) - e_\phi(s',a')\|_H + V^*(s',g)]
\label{eq:bellman_quasi}
\end{equation}
subject to $\|e_\phi(s,a) - e_\phi(s',a')\|_H \leq \delta$.
Now, consider the graph $G$ where each node represents a state-action pair $(s,a)$, and edges represent transitions with weights $-V^*(s,s')$. The minimum-cost path in this graph from any state $s$ to the goal $g$ will correspond to the sequence of actions that maximizes $V^*(s,g)$ while respecting the safety constraints at each step. The policy chooses the action $a$ that maximizes the right-hand side of \eqref{eq:bellman_quasi}. This is equivalent to choosing the edge in $G$ with the minimum weight (since the weights are negative values of $V^*$), subject to the constraint $\|e_\phi(s,a) - e_\phi(s',a')\|_H \leq \delta$. Therefore, the minimum-cost path in $G$, constructed using the learned quasimetric embedding $e_\phi(s,a)$, yields the optimal constrained policy $\pi^*$ that maximizes $V^*(s,g)$ while satisfying the safety constraints at each step of the trajectory.
\end{proof}

\section{Experiments and Results}


\begin{figure*}[!htb]
    \centering
    \subfloat[]{%
        \includegraphics[width=0.32\textwidth,height=3cm]{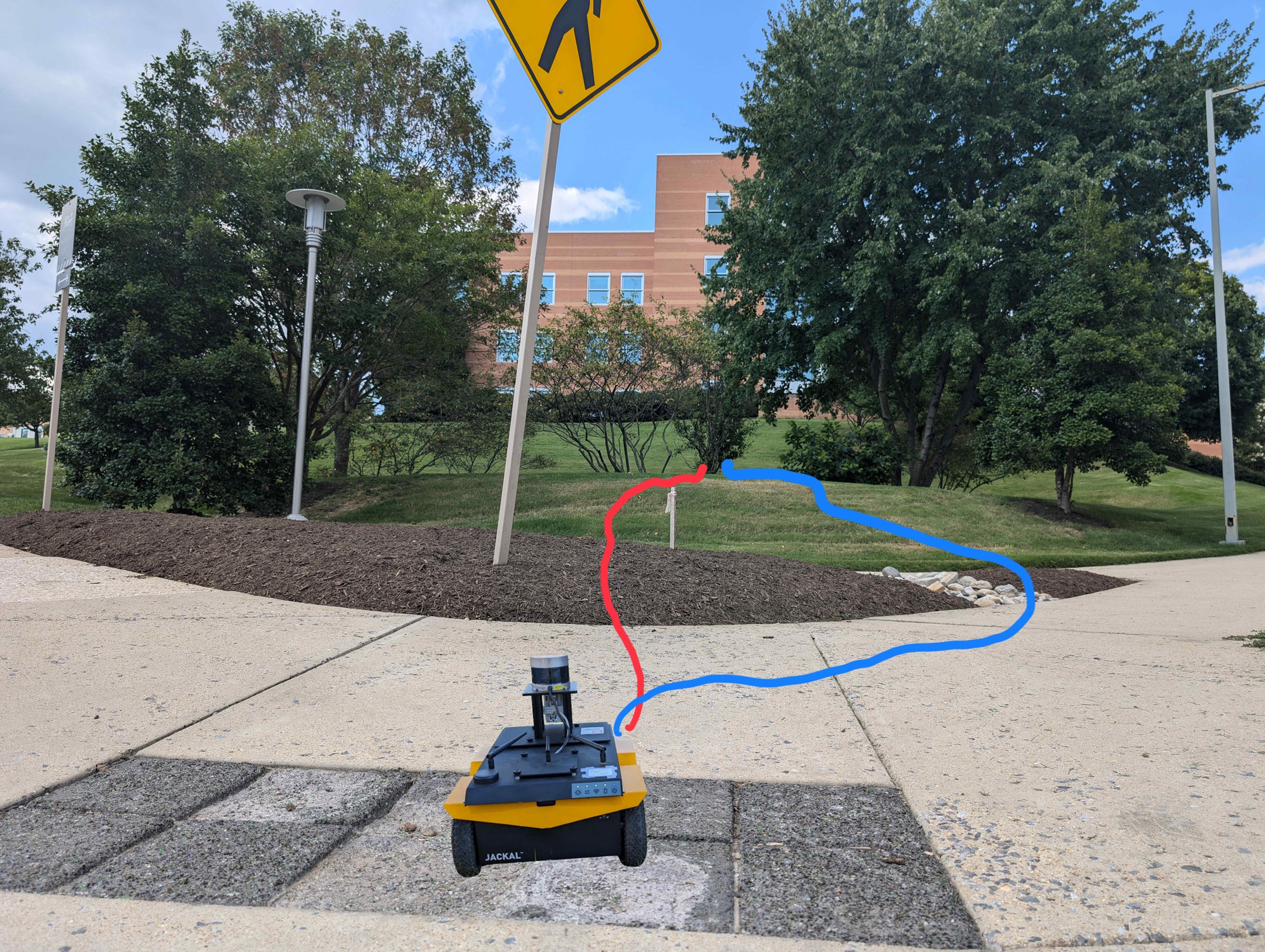}
    }
    \hfill 
    \subfloat[]{%
        \includegraphics[width=0.32\textwidth,height=3cm]{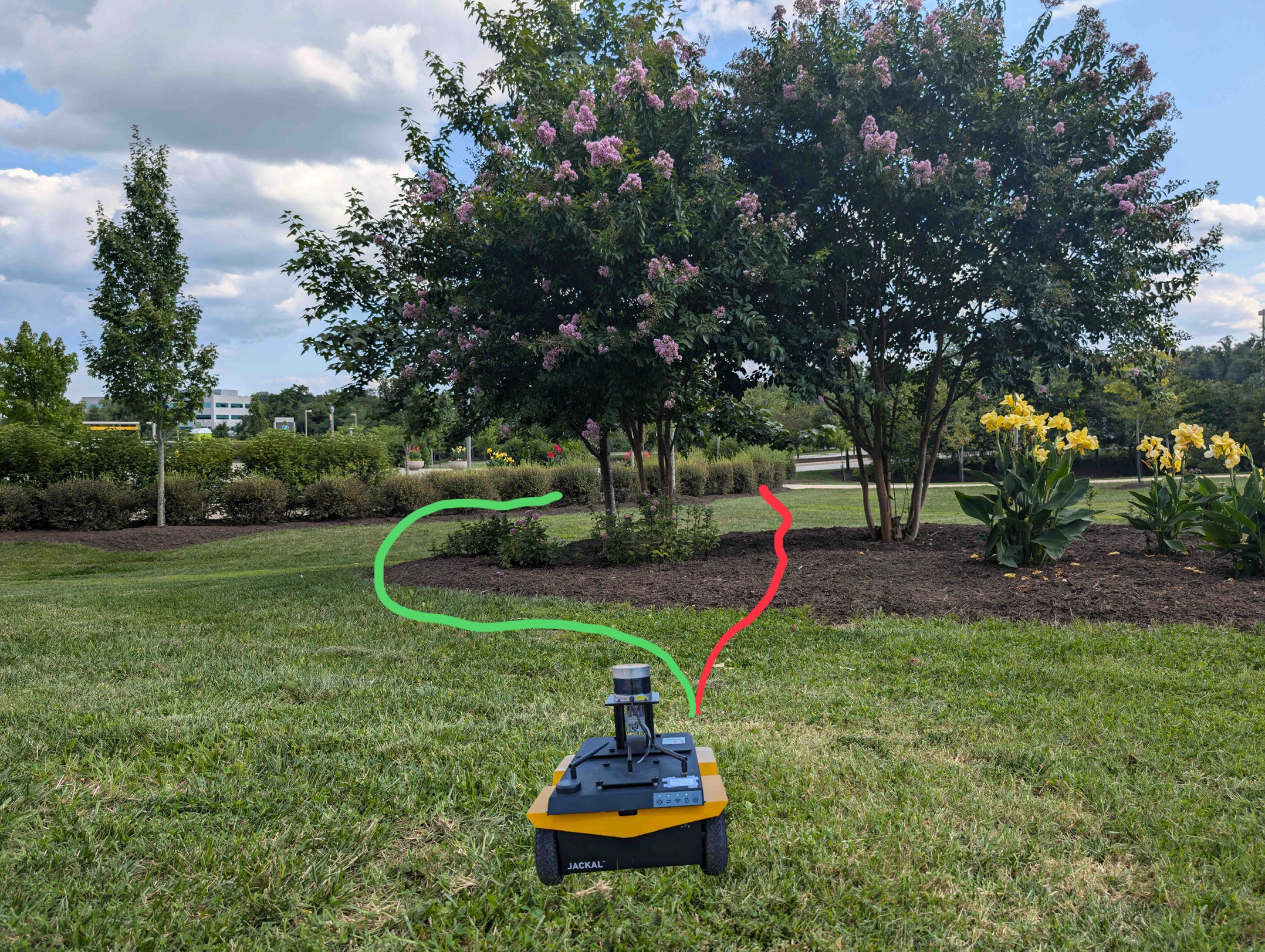}
    }
    \hfill
    \subfloat[]{%
        \includegraphics[width=0.32\textwidth,height=3cm]{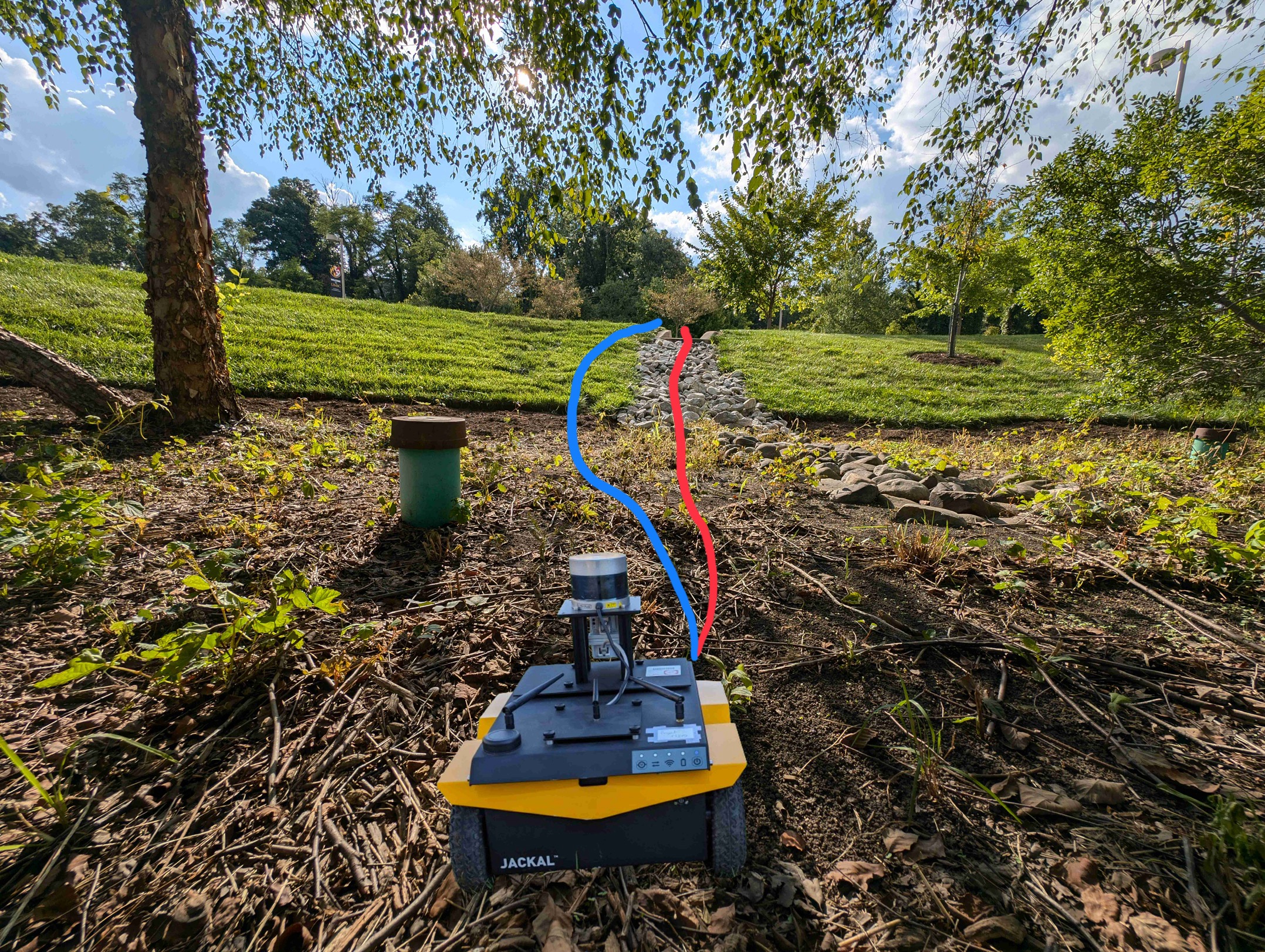}
    }

\caption{Real-world navigation experiments using \textit{QuasiNav} on a Clearpath Jackal robot, demonstrating its ability to optimize navigation in complex outdoor environments with asymmetric traversal costs: (a) Undulating Terrain Navigation: The robot selects a longer, gentler path (blue) over a steeper alternative (red), prioritizing energy efficiency. (b) Hill Traversal with Asymmetric Costs: \textit{QuasiNav} chooses a winding, lower-gradient route (blue) instead of a direct steep climb (red), reflecting its understanding of asymmetric elevation costs. (c) Directionally Dependent Terrain Traversal: The robot opts for a longer but less challenging path (blue), avoiding a direct but energy-intensive route (red) through difficult terrain. These experiments validate \textit{QuasiNav}'s effectiveness in real-world scenarios, successfully translating learned policies from simulation to complex, unstructured outdoor settings.}
    \label{fig:real_world_nav}
\end{figure*}

\subsection{Navigation Scenarios and Reward Design}

We evaluate \textit{QuasiNav} across three challenging navigation scenarios, designed to test its ability to handle asymmetric costs and safety constraints. The reward functions are formulated to penalize energy-inefficient, unsafe, or slow actions.

\noindent \textbf{Scenario 1 (Undulating Terrain Navigation) : } In this scenario, the robot navigates an environment with varying slopes and soil types. The reward function balances energy efficiency with time:
\begin{equation}
r(s,a,s') = -(\alpha_\text{energy} \cdot C_\text{energy}(s,a,s') + \lambda_\text{time} \cdot \Delta t)
\label{eq:reward_undulating}
\end{equation}
where $C_\text{energy}(s,a,s')$ is the energy cost:
\begin{equation}
C_\text{energy}(s,a,s') = \frac{\Delta d}{\cos(\theta)} \cdot f_\text{terrain}(s) \cdot (1 + \beta_\text{slope} \cdot \Delta z)
\label{eq:cost_energy}
\end{equation}
Here, $\theta$ is the slope angle, $\Delta d$ is the distance traveled, $\Delta z$ is the elevation change, $f_\text{terrain}(s)$ is the terrain friction coefficient, and $\Delta t$ is the time taken for the transition. This reward function captures the trade-off between taking longer, easier paths and shorter, more challenging ones.

\noindent \textbf{Scenario 2 (Asymmetric Hill Traversal) : } This scenario tests the robot's ability to handle asymmetric uphill and downhill traversal costs. The reward function emphasizes the asymmetric nature of hill traversal:
\begin{equation}
r(s,a,s') = -(\alpha_\text{hill} \cdot C_\text{hill}(s,a,s') + \lambda_\text{safety} \cdot I_\text{unsafe}(s,a))
\label{eq:reward_hill}
\end{equation}
where $C_\text{hill}(s,a,s')$ is modeled based on the elevation change $\Delta z$:
\begin{equation}
C_\text{hill}(s,a,s') = 
\begin{cases}
\kappa_\text{up} \cdot \Delta z \cdot f_\text{terrain}(s) & \text{if } \Delta z > 0 \text{ (uphill)} \\
\kappa_\text{down} \cdot |\Delta z| \cdot f_\text{terrain}(s) & \text{if } \Delta z \leq 0 \text{ (downhill)}
\end{cases}
\label{eq:cost_hill}
\end{equation}
and $I_\text{unsafe}(s,a)$ is an indicator function that applies a penalty if the robot attempts to traverse unsafe slopes. This reward function penalizes uphill movement more heavily and incorporates a safety penalty for risky actions.

\noindent \textbf{Scenario 3 (Directionally Dependent Terrain Traversal) : } In this scenario, the cost of traversal depends on the direction of travel. The reward function captures these directional dependencies:
\begin{equation}
r(s,a,s') = -(C_\text{terrain}(s,a,s') \cdot f_\text{direction}(s,a) + \lambda_\text{friction} \cdot f_\text{terrain}(s))
\label{eq:reward_directional}
\end{equation}
where $C_\text{terrain}(s,a,s')$ is the base cost of traveling between $s$ and $s'$:
\begin{equation}
C_\text{terrain}(s,a,s') = \frac{\Delta d}{v(s)} \cdot (1 + \beta_\text{direction} \cdot \Delta z)
\label{eq:cost_terrain}
\end{equation}
and $f_\text{direction}(s, a)$ is a scaling factor that depends on the direction of traversal relative to the terrain. This reward function accounts for how different terrains may be easier or harder to navigate based on the direction of travel.

In all scenarios, the reward functions are designed to penalize energy-inefficient and unsafe actions, encouraging the robot to optimize for safe and efficient traversal. The overall goal is to maximize the cumulative reward:

\begin{equation}
R = \sum_{t=0}^T \gamma^t r(s_t, a_t, s_{t+1})
\label{eq:cumulative_reward}
\end{equation}
where $\gamma$ is the discount factor that prioritizes short-term rewards over long-term gains.

\subsection{Energy Consumption Model and Measurement}

To accurately assess navigation efficiency, we developed a comprehensive energy consumption model that accounts for terrain-specific factors. The instantaneous power consumption $P(t)$ at time $t$ is modeled as:
\begin{equation}
P(t) = P_\text{base} + P_\text{motion}(v(t), \alpha(t)) + P_\text{terrain}(f(t), \theta(t))
\end{equation}
where $P_\text{base}$ is the base power consumption of the robot, $P_\text{motion}$ is the power required for motion as a function of velocity $v(t)$ and acceleration $\alpha(t)$, and $P_\text{terrain}$ is the additional power required to overcome terrain effects, dependent on the friction coefficient $f(t)$ and slope angle $\theta(t)$. In our experiments, we used an onboard power monitoring system to measure the instantaneous current draw and voltage from the robot's battery. The total energy consumption $E$ for a trajectory of duration $T$ is calculated as:
\begin{equation}
E = \int_0^T P(t) dt
\end{equation}
This energy measurement is incorporated into our reward function (Equations \ref{eq:reward_undulating}, \ref{eq:reward_hill}, and \ref{eq:reward_directional}) and serves as a key evaluation metric for comparing the efficiency of different navigation strategies.
\subsection{Implementation}

\textit{QuasiNav} was implemented in PyTorch and trained using simulated terrains, leveraging ROS Noetic and Unity simulation framework for realistic outdoor environments. The training setup included a Clearpath Husky robot equipped with a Velodyne 3D LiDAR in the simulation, and the network was trained on a workstation with an Intel Core i9 CPU and NVIDIA RTX 3090 GPU. The model used a batch size of 128 and the Adam optimizer with a learning rate of $10^{-4}$, training over 1 million steps. For real-time deployment, a Clearpath Jackal UGV with a Velodyne VLP-32C LiDAR and a fixed IP camera was used, featuring onboard computing with an Intel i7 CPU and NVIDIA GTX 1650 Ti GPU. The LiDAR provided 3D point cloud data for obstacle detection and avoidance, incorporating techniques like the Dynamic Window Approach (DWA)~\cite{fox1997dynamic} for navigation.

\subsection{Evaluation Metrics}

The performance of \textit{QuasiNav} was assessed using several key metrics that reflect its efficiency, safety, and adaptability in unstructured outdoor environments. The primary metrics include:

\begin{itemize}
    \item \textbf{Success Rate:} The percentage of trials where the robot successfully reached the goal while adhering to safety constraints.
    \item \textbf{Energy Consumption:} Estimated based on the traversal costs related to elevation changes and roughness, this metric evaluates how well \textit{QuasiNav} optimizes for energy-efficient navigation.
    \item \textbf{Safety Compliance:} Evaluated by the frequency of violations of safety constraints, such as navigating through restricted zones or exceeding allowed inclinations.
    \item \textbf{Asymmetric Cost Handling:} Assessed by comparing the traversal costs of uphill versus downhill paths, demonstrating \textit{QuasiNav}'s capability to account for directionally dependent costs.
    
\end{itemize}

\subsection{Performance Comparison and Analysis}

To evaluate \textit{QuasiNav}, we compared it against state-of-the-art algorithms, including QRL ~\cite{tongzhouw2023qrl}, HTRON ~\cite{weerakoon2022htron}, TERP ~\cite{weerakoon2022terp} and GANav ~\cite{guan2022ga}. The evaluation was conducted across three scenarios: undulating terrain, hill traversal with asymmetric costs, and directionally dependent terrain traversal. Table \ref{tab:performance_comparison} summarizes the results, demonstrating \textit{QuasiNav}’s superior performance across key metrics such as success rate, energy efficiency, safety compliance, and asymmetric cost handling.

\begin{table}[ht]
\centering
\scriptsize
\begin{tabular}{|l|c|c|c|c|}
\hline
\textbf{Metrics} & \textbf{Methods} & \textbf{Scenario 1} & \textbf{Scenario 2} & \textbf{Scenario 3} \\
                 &                  & \textbf{Undulating}  & \textbf{Hill}        & \textbf{Directional}  \\
                 &                  & \textbf{Terrain}     & \textbf{Traversal}   & \textbf{Terrain}      \\
\hline
\textbf{Success} & QuasiNav (Ours)  & \textbf{94.0}        & \textbf{93.2}        & \textbf{92.5}         \\
\textbf{Rate}    & QRL              & 90.5                 & 88.4                 & 89.0                  \\
\textbf{(\%)}    & TERP             & 89.0                 & 87.5                 & 88.2                  \\
                 & HTRON            & 86.2                 & 84.5                 & 85.3                  \\
                 & GANav            & 85.0                 & 83.5                 & 84.0                  \\
\hline
\textbf{Energy}  & QuasiNav (Ours)  & \textbf{140.0}       & \textbf{145.2}       & \textbf{142.6}        \\
\textbf{Eff.}    & QRL              & 150.5                & 155.0                & 152.0                 \\
\textbf{(J/m)}   & TERP             & 155.0                & 159.5                & 157.0                 \\
                 & HTRON            & 160.3                & 164.5                & 162.1                 \\
                 & GANav            & 145.0                & 150.0                & 148.0                 \\
\hline
\textbf{Safety}  & QuasiNav (Ours)  & \textbf{1.5}         & \textbf{2.0}         & \textbf{2.1}          \\
\textbf{Viol.}   & QRL              & 4.0                  & 3.7                  & 4.2                   \\
\textbf{(\%)}    & TERP             & 3.0                  & 3.5                  & 3.2                   \\
                 & HTRON            & 5.2                  & 4.5                  & 5.0                   \\
                 & GANav            & 3.5                  & 4.0                  & 3.8                   \\
\hline
\textbf{Asym.}   & QuasiNav (Ours)  & \textbf{Excellent}   & \textbf{Excellent}   & \textbf{Excellent}    \\
\textbf{Cost}    & QRL              & Good                 & Good                 & Good                  \\
\textbf{Handling}& TERP             & Good                 & Good                 & Good                  \\
                 & HTRON            & Fair                 & Fair                 & Fair                  \\
                 & GANav            & Fair                 & Fair                 & Fair                  \\
\hline
\end{tabular}
\caption{Performance Comparison of \textit{QuasiNav} and Baseline Algorithms Across Scenarios}
\label{tab:performance_comparison}
\end{table}

The results show that \textit{QuasiNav} outperforms QRL, HTRON, TERP, and GANav across all key metrics. \textit{QuasiNav} achieved the highest success rate of 94\%, demonstrating its ability to navigate complex terrains efficiently while maintaining high safety compliance. TERP, which is specifically designed for elevation-aware navigation, performs better than HTRON in our scenarios, reflecting its enhanced ability to handle undulating and asymmetric terrain. However, TERP still falls short of \textit{QuasiNav}, particularly in scenarios that require real-time adaptation to asymmetric traversal costs.

\textbf{Asymmetric Cost Handling}: This metric assesses each method's ability to manage directional dependencies in traversal costs, such as preferring downhill paths when energetically favorable or safer. \textit{QuasiNav} excels in this area due to its integration of quasimetric embeddings that accurately model these asymmetries, while TERP and GANav offer good performance with their elevation and terrain-awareness features. HTRON, however, shows limited capacity to handle asymmetric costs effectively, leading to less optimal path choices in complex terrains.

\subsection{Ablation Studies}
To evaluate the individual contributions of quasimetric embeddings and adaptive constraint tightening to \textit{QuasiNav}’s overall performance, we conducted ablation studies. Table \ref{tab:ablation} summarizes the results, illustrating the impact of removing these components. The full \textit{QuasiNav} configuration achieved the highest success rate, energy efficiency, and lowest safety violations, highlighting the importance of both quasimetric embeddings and adaptive constraint tightening. The absence of quasimetric embeddings led to significant reductions in success rate and energy efficiency, while omitting adaptive constraint tightening resulted in increased safety violations.

\begin{table}[h]
\centering
\scriptsize
\begin{tabular}{|l|c|c|c|}
\hline
\textbf{Configuration} & \textbf{Success} & \textbf{Energy} & \textbf{Safety} \\
                       & \textbf{Rate (\%)} & \textbf{Efficiency (J/m)} & \textbf{Violations (\%)} \\
\hline
QuasiNav (Full)        & \textbf{94.7}      & \textbf{142.6}           & \textbf{2.1} \\
\hline
w/o Quasimetric        & 85.3               & 161.5                     & 5.7 \\
Embedding              &                    &                           &     \\
\hline
w/o Adaptive           & 89.1               & 145.3                     & 4.2 \\
Constraint Tightening  &                    &                           &     \\
\hline
\end{tabular}
\caption{Ablation Study Results}
\label{tab:ablation}
\end{table}

\subsection{Complexity Analysis}

QuasiNav's computational complexity is primarily determined by the quasimetric embedding, policy optimization, and adaptive constraint tightening processes. The time complexity per iteration is $\mathcal{O}(d_e + |S||A| + k)$, where $d_e$ is the embedding dimension, $|S|$ and $|A|$ are the sizes of the state and action spaces respectively, and $k$ is the number of constraints. The space complexity is $\mathcal{O}(d_e + |S||A| + N)$, where $N$ is the replay buffer size. In practice, $d_e$ and $k$ are typically much smaller than $|S||A|$. Empirically, QuasiNav maintains real-time performance ($>10$ Hz) on our hardware setup for environments with up to $10^6$ states, with runtime scaling linearly with environment complexity.



\section{Conclusion, Limitations, and Future Directions}
\label{sec:conclusion}

In this paper, we presented \textit{QuasiNav}, an innovative approach to safe and efficient navigation in complex, unstructured outdoor environments using Quasimetric Reinforcement Learning (QRL) combined with constrained policy optimization. \textit{QuasiNav}’s key features include quasimetric embeddings to accurately capture asymmetric traversal costs and adaptive constraint tightening to ensure safety during navigation. Our evaluation across various scenarios demonstrated \textit{QuasiNav}’s superior performance in achieving higher success rates, better energy efficiency, and improved safety compliance compared to the baselines. While \textit{QuasiNav} handles most scenarios effectively, our analysis of failure cases identified challenges in extremely complex terrains and highly dynamic conditions. Future work will focus on enhancing safety mechanisms, refining adaptive embeddings, and improving sim-to-real transfer techniques to further advance \textit{QuasiNav}'s robustness and applicability in diverse navigation tasks.

\bibliographystyle{unsrt}
\bibliography{bibliography}
\end{document}